\PassOptionsToPackage{warn}{textcomp}
\documentclass[a4paper]{article}
\usepackage{INTERSPEECH2018}
\usepackage[utf8]{inputenc}
\usepackage[T1]{fontenc}
\usepackage{microtype}
\usepackage[hidelinks]{hyperref}
\usepackage{stmaryrd} 
\usepackage{comment}
\usepackage{todonotes}
\usepackage{relsize}
\usepackage{subcaption}

\newcommand{\uprarrow}{\raisebox{.1ex}{\rotatebox[origin=c]{45}{$\shortrightarrow$}}}
\newcommand{\cmp}{\textasciitilde}
\newcommand{\mysiggg}{\textbf{\textasteriskcentered{}\textasteriskcentered{}\textasteriskcentered}}
\newcommand{\mysigg}{\textbf{\textasteriskcentered{}\textasteriskcentered}}
\newcommand{\mysig}{\textbf{\textasteriskcentered}}
\newcommand{\myinsig}{\textsc{ns}}

\title{An Empirical Analysis of the Correlation of Syntax and Prosody}
\name{Arne Köhn$^1$, Timo Baumann$^2$, Oskar Dörfler$^1$}
\address{
  $^1$Natural Language Systems Group, Department of Informatics, Universität Hamburg, Germany\\
  $^2$Language Technology Institute, Carnegie Mellon University, Pittsburgh, USA}
\email{\{koehn,3doerfle\}@informatik.uni-hamburg.de, tbaumann@cs.cmu.edu}

\makeatletter
\hypersetup{
  pdfinfo={
    Title={\@title},
    Author={Arne Köhn, Timo Baumann, Oskar Dörfler},
  }
}
\makeatother

\begin{document}

\maketitle
\begin{abstract}
The relation of syntax and prosody (the syntax--prosody interface) 
has been an active area of research, mostly in linguistics and typically 
studied under controlled conditions. 
More recently, prosody has also been successfully used in the data-based 
training of syntax parsers. 
However, there is a gap between the controlled and detailed study
 of the individual effects between syntax and prosody 
and the large-scale application of prosody in syntactic parsing with only 
a shallow analysis of the respective influences.
In this paper, we close the gap by investigating the significance of 
correlations of prosodic realization with specific syntactic functions
using linear mixed effects models
in a very large corpus of read-out German encyclopedic texts.
Using this corpus, we are able to analyze prosodic structuring
performed by a diverse set of speakers while they try to optimize
factual content delivery. 
After normalization by speaker, we obtain significant effects, 
e.\,g.\ confirming that the subject function, as compared to the object function,
has a positive effect on pitch and duration of a word, but a negative effect on loudness.
%

\end{abstract}
 
\section{Introduction}

Spoken language features two structural systems -- syntax on the language side 
and prosody on the speech side -- with a strong interrelation that has 
been a long-running topic of scientific research. 
Our contribution is to bring together large-scale corpus-based analysis using 
automatic tools with the breakdown of individual effects of the interplay to 
underpin linguistic insight.

Looking at pausing and constituency, it was first observed that 
most pauses appear at major syntactic constituent breaks \cite{grosjean1975analyse}, 
and that pause duration relates to the strength of breaks 
between constituents of a read sentence \cite{GROSJEAN197958}.
Using controlled experiments rather than observations, 
Price et al.\ \cite{priceetal1991} found that 
syntactic ambiguities can be recovered by listeners from the prosodic realization, 
and remark that this happens 
``primarily based on boundary phenomena, although prominences sometimes play a role'' \cite{priceetal1991}.
In a similar way, Beach \cite{BEACH1991644} shows that prosody in a partial sentence 
can be used to (partially) predict sentence structure, 
in particular subject/object opposition, 
and using synthesized speech (thus limiting other influences).
Weber et al.\ ``conclude that in addition to manipulating attachment ambiguities,
prosody can influence the interpretation of constituent order ambiguities'' \cite{WEBER2006B63}, 
hence pointing towards much broader influences, 
using eye-tracking in laboratory phonetics settings. 
Neurolinguists, likewise, have found evidence for the integrated processing 
of prosody and syntax in the human brain \cite{doi:10.1162/jocn.2006.18.10.1696}, 
in particular of intonation and sentence finality.
Thus, we conclude that there is ample evidence from detailed analysis and 
controlled studies on the influences between prosody and syntax and the merit
of prosody to differentiate syntactic ambiguities. In particular, prosody helps 
to differentiate constituent boundaries (in particular through pausing), 
attachment ambiguities, and syntactic function (potentially through features
more closely related to prominence like duration, loudness and pitch).

The merit of prosody has indeed been used to improve syntax parsing, by
including cues into PCFG parsers \cite{gregory2004sentence,kahn2005effective} 
and very recently in neural networks-based dependency parsing \cite{N18-1007}. 
While the addition of discrete prosodic symbols has been helpful in 
tasks like speech recognition and understanding \cite{shriberg2004prosody},
this was not helpful for parsing \cite{gregory2004sentence}. 
In contrast,
a neural networks-based syntax parser can use a continuous representation of 
prosody and shows a merit on overall parsing accuracy 
for pausing, word duration, loudness and pitch \cite{N18-1007}.

The work on acoustic parsing, while based on large corpora of syntactically 
annotated language has two limitations: 
(a) The resulting parser models cannot easily be analyzed in terms of which 
prosodic aspects help towards resolving what syntactic ambiguity. 
The models merely show that overall, prosody helps to improve parsing performance
but do not explain concretely which syntactic structures correlate with which 
prosodic properties. 
(b) The work has only been performed on conversational language \cite{calhoun2010nxt}
and prosody was shown to be particularly helpful for parsing sentences with disfluencies.
This leaves open the question of whether prosody would also help for parsing 
more canonical, read language which does not contain such disfluencies, or whether 
the positive influence of prosody is largely valuable to recover from disfluencies.

We analyze, in a large and diverse corpus of read German encyclopaedic material 
as collected from the Spoken Wikipedia\footnote{%
\url{http://en.wikipedia.org/wiki/Wikipedia:WikiProject_Spoken_Wikipedia};
also contains links to other languages.}~\cite{swcLRE2018}, 
the relation of syntactic dependency structure and prosodic realization
using linear mixed effects models \cite{brauer2017linear}. 
We analyze the correlation of prosodic features measured on a word and the 
syntactic function assigned to that word (or the word's head). 
We select the conditions to test based on linguistic intuition and find 
significant results for a number of oppositions (such as whether a noun is used 
as a subject or object).
Similar in spirit to our work,
prosodic features and their relation to broad syntactic features such as 
content vs.\ function words are analyzed in \cite{heldner2003exploring}.
Although that work uses a small corpus of conversations, they similarly use 
automatic alignments and automatic annotations (in that case parts of speech).

The remainder of this paper is structured as follows: in Section~\ref{sec:model},
we explain in detail the method of using a linear mixed effects model for 
finding significant differences in prosodic characteristics of words with different 
syntactic functions. In Section~\ref{sec:data}, we describe the data that we use 
in the experiment which we describe and evaluate in Section~\ref{sec:experiment}.
We discuss our results and conclude in Section~\ref{sec:discussion}.

\begin{figure*}
  \begin{subfigure}[b]{.5\linewidth}
    \includegraphics[scale=0.6]{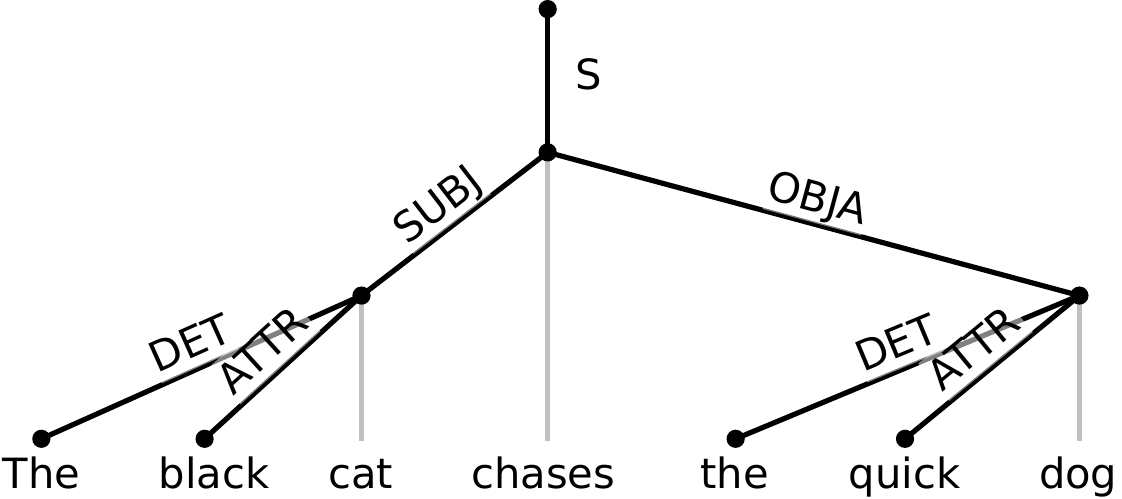}
    \subcaption{Example for noun-related structures (\emph{subj} and \emph{obja})}\label{fig:idea-a}
  \end{subfigure}
  \begin{subfigure}[b]{.5\linewidth}
    \includegraphics[scale=0.6]{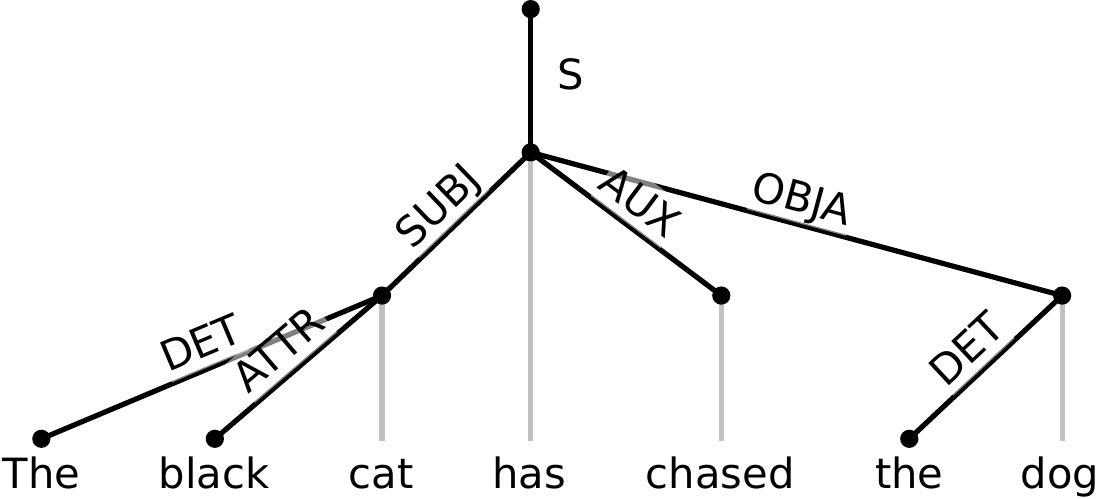}
    \subcaption{Example for an \emph{aux} function}\label{fig:idea-b}
  \end{subfigure}
  \begin{subfigure}[b]{.5\linewidth}
    \includegraphics[scale=0.6]{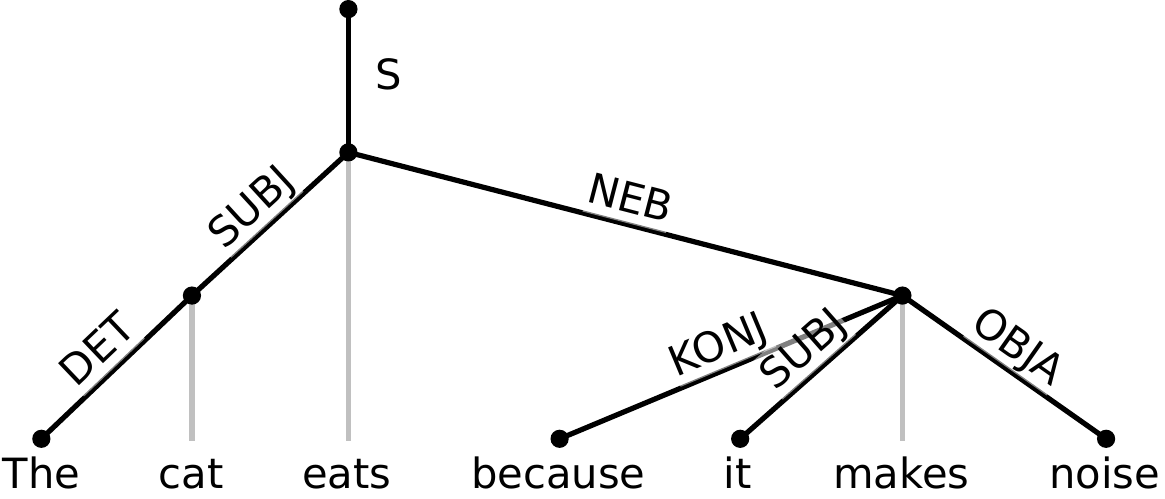}
    \subcaption{Example for a subordinate clause (\emph{neb})}\label{fig:idea-c}
  \end{subfigure}
  \begin{subfigure}[b]{.5\linewidth}
    \includegraphics[scale=0.6]{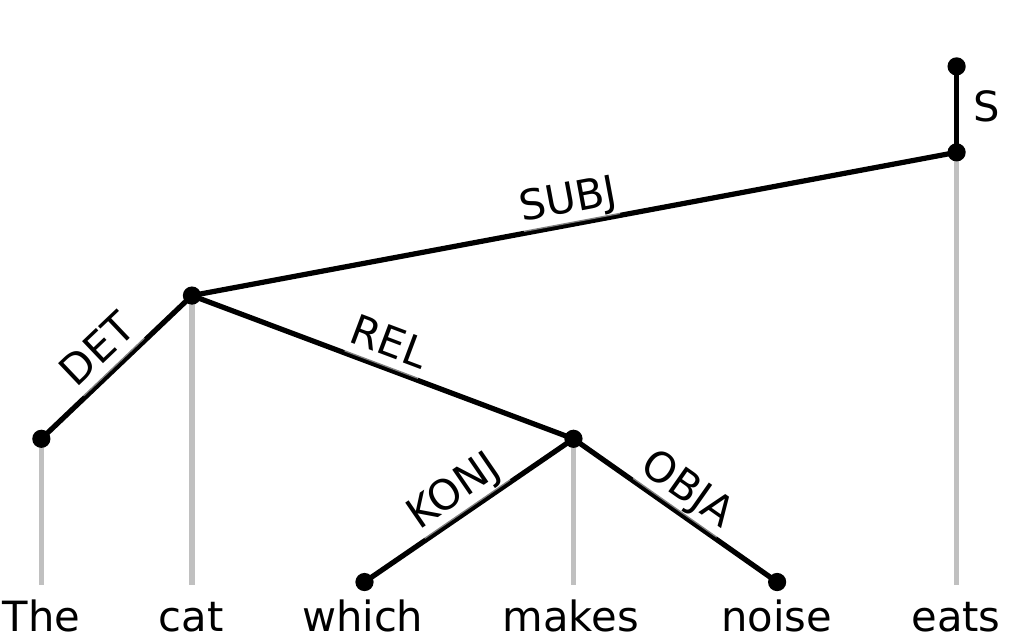}
    \subcaption{Example for a relative clause (\emph{rel})}\label{fig:idea-d}
\end{subfigure}
\caption{%
Dependency trees exemplifying the syntactic structures we use to predict prosodic features.
  \label{fig:idea}%
}
\end{figure*}

\section{Method and Statistical Model}
\label{sec:model}


We focus our analysis of the syntax--prosody interface on finding (significant) 
correlations between syntactic functions of words and prosodic features of those words.
We ignore the question of causality in this paper (and given the neurolinguistic 
research cited above there probably is no clear causality but a coordinated interplay). 
Also, we focus on very simple features that can be automatically extracted: 
tempo, avg.\ pitch, avg.\ loudness and the duration of a following pause. 
These features can be measured easily and objectively and do not take into account 
any linguistic structure (unlike, e.\,g.\ \cite{rosenberg2010autobi}).

We find correlations by performing a likelihood ratio test between two linear models: 
(a) the basic model is fitted to predict the outcome, 
e.\,g.\ the mean pitch of the word using non-lexicalized textual features
such as length of the word and position in the sentence.
(b) The extended model performs the same task but uses the syntactic function
as an additional feature.  
For each model (a) and (b) we compute how well it fits the data
(i.\,e.\ compute the likelihood of the model).
If the extended model fits the data better than the
basic model, the syntactic function yields a benefit in predicting prosody 
even after accounting for the features contained in the basic model.  
This benefit can be expressed by dividing the likelihood of the basic model 
by the likelihood of the extended model and this likelihood ratio is
used to compute \emph{p}-values using the likelihood ratio test.

Different speakers perform differently, and recordings
of the same speaker might differ.\footnote{Some speakers recorded over
  a span of ten years.}
To account for this, we take into account speaker and recording ID and
each linear model is built using fixed effects
(coefficients are shared between articles) and random effects (coefficients are
optimized for each article to capture speaker characteristics),
which yields a linear mixed effects model.  
We use the features themselves as predictors, 
as well as the interactions between features (i.\,e.\ feature combinations).
The extended models only add one binary categorial feature each, 
namely to which of two syntactic groups an example belongs.
To obtain a magnitude of change between the two groups, the
coefficients from the basic model are used in the extended model and
only the coefficient for the syntax information is fitted.  This
coefficient is the average difference between the two groups after
controlling for the features of the basic model and is reported as effect in Cent/dB/ms in
Table~\ref{tab:results}.
Hypothesis testing is performed using 
the \texttt{lme4} R package \cite{bates-etal-2014-lme4}; for an in-depth discussion
of hypothesis testing with linear mixed effects models see e.\,g.\ \cite{brauer2017linear}.


We perform two types of tests:  
The first type of test compares words that could fill the same syntactic roles
using their function in the sentence as a predictor
for prosodic features.  For example, nouns and pronouns can act both as object (\emph{obja}) and
as subject (\emph{subj}) in a sentence (see ``cat'' and ``dog'' in the example in
Figure~\ref{fig:idea-a}).  To shorten the notation, we write
A\cmp B to denote a comparison of words in context A to words
in context B (in this case, subj\cmp obja).
The second type of test compares words
that fill the same syntactic function 
but are attached to words that fill different functions.  As
an example, the word ``the'' occurs twice in the example in
Figure~\ref{fig:idea-a}, once as a determiner (\emph{det}) for the subject and once
as a determiner for the object.  We write A\uprarrow(B\cmp C) to
denote a comparison of words having dependency label A attached to
words filling role B with  words having dependency label A attached to
words filling role C (in this case, det\uprarrow(subj\cmp obja)).

Our comparisons only entail pairs of syntactic roles where words that can fill
one syntactic role can also fill the other, e.\,g.\ the subject and object roles
can both be filled by nouns, proper nouns, names, or pronouns.
We could e.\,g.\ test for
subj\cmp det, but due to the two groups being disjoint, there would be
no meaningful interpretation of the results and the fitted
models might just pick up systematic differences between
nouns and determiners, via their dependency labels.

Our approach could be seen as needlessly complicated; a
straightforward (but wrong) way of testing the effect of syntactic
structure on prosodic features would be to extract pairs of syntax and
prosody features and determine the correlation between both.  
However, this
approach would yield spurious significance as important mediating
factors would not be modeled. For example, a difference in pitch between
subjects and objects might simply stem from the fact that objects tend
to occur later in a sentence than objects and pitch tends to
decrease over the sentence as subglottal pressure decreases \cite{gelfer1987}.  
We would not really measure the influence of syntax 
but merely the influence of the word's position, expressed
through the correlation with syntax.







\section{Data and Setup}
\label{sec:data}

\begin{table*}
  \caption{Experimental settings, their significance level and the coefficients of the added syntactic predictors
  in the extended model}
  \label{tab:results}
  \centering
\setlength{\tabcolsep}{9pt}
\begin{tabular}{@{}lrlllllllll@{}}
  \toprule
        &       &              & det\uprarrow& attr\uprarrow&           &           &             &\multicolumn{3}{c}{aux\uprarrow\ldots} \\
                                    \cmidrule(lr){4-5}                                               \cmidrule(lr){9-11}
        &       & subj\cmp obja&  \multicolumn{2}{c}{\ldots\uprarrow(subj\cmp obja)} & s\cmp neb & s\cmp rel & s\cmp aux  & (s\cmp neb) & (s\cmp rel)  & (s\cmp aux) \\
\midrule
  pitch	&p-value& \mysiggg & \myinsig & \mysigg  & \myinsig & \mysiggg & \myinsig & \myinsig & \mysiggg & \mysiggg \\
& effect in Cent& 19.38    & ---      & 25.14    & ---      & 39.38    & ---      & ---      & 22.81    & 53.77    \\[.5em]
power	&p-value& \mysiggg & 0.06     & \mysig   & \mysig   & \mysiggg & \myinsig & \myinsig & \myinsig & \myinsig \\
& effect in dB	& -0.01    &0.005     & 0.01     &  0.01    & 0.02     & ---      & ---      & ---      & ---      \\[.5em]
duration&p-value& \mysiggg &\mysiggg  & \myinsig & \mysiggg & \mysiggg & \mysiggg & 0.12     & \mysiggg & \mysig   \\
& effect in ms	& 35.42    & -14.62   & ---      & -17.8    & -23.03   & -29.32   & 15.88    & 19.03    & 30.04    \\[.5em]
pause	&p-value& \mysig   &\mysigg   & \myinsig & \myinsig & \mysig   & \mysiggg & \myinsig & \mysiggg & \myinsig \\
& effect in ms	& 3.81     & 6.14     & ---      & ---      & -5.52    & -10.17   & ---      & 10.73    & ---      \\          
  \bottomrule
\end{tabular}

{\vspace{2pt}\smaller Significance levels: \mysiggg: $< .001$, \mysigg: $<.01$, \mysig: $<.05$; clearly non-significant results: `\myinsig'.  Effect is positive iff first > last, e.\,g.\  pitch higher for \emph{subj} than \emph{obj}.
}
\end{table*}

We use recordings from the Spoken Wikipedia as a sample of read 
\emph{speech in the wild} rather than laboratory-collected speech samples.
The Spoken Wikipedia project unites volunteer readers who devote significant 
amounts of time and effort into producing read versions of Wikipedia articles 
read by a broad speaker population.
It can thus be considered a valid source of speech produced by ambitious but 
not always perfect readers. The large number of readers make the corpus a valid 
sample of speech (although gender is imbalanced with only \texttildelow{}10\,\% female readers).

The data has been prepared as a corpus \cite{swcLRE2018}.\footnote{\label{fnSWC}%
Available at: \url{http://islrn.org/resources/684-927-624-257-3/} 
and \url{http://nats.gitlab.io/swc}.}
Importantly, the annotation includes a linguistic sentence segmentation and 
tokenization and the relation of original and normalized text has been preserved, 
allowing the timings of the aligned normalized text to be mapped to each original 
text token, thus bridging the gap between speech and language processing.

We limit our analysis to the German sub-corpus of the Spoken Wikipedia which contains 
some 1000 articles totaling 386\,h of audio (360\,h after VAD) and 3\,M word tokens
read by 350 different speakers \cite{swcLRE2018}.
The alignment favors quality over coverage and hence only about 70\,\% of the 
word tokens have alignment information available. 
For simplicity, we limit our analysis to those sentences in which every word 
has alignment information which yields 31,803 fully aligned sentences\footnote{%
The published version of the corpus, unlike the pre-release that we use here, 
contains about twice as many fully aligned sentences.}
with a total of 348,062 word tokens in 57,265 word forms.

We enrich the annotation with a dependency tree as well as Part-of-Speech tags
for each sentence in our dataset using TurboParser
\cite{martins-almeida-smith:2013:Short} trained for German 
on the Hamburg Dependency Treebank \cite{Foth14}.
The reported performance of the parser is $>$93\,\% labeled accuracy \cite{Foth14}.
We parsed the complete German Spoken Wikipedia and these parses as well as all other code and data
needed to reproduce our findings are freely available\footnotemark[3].

Our 32k fully aligned sentences span 46 hours of speech.
We extract two types of data:  textual features to predict prosodic features, and the
prosodic ground truths to be predicted.
Regarding textual features, we compute, for every word in our dataset, the 
canonical word duration using the duration predictor by Hal Daumé III\footnote{%
  See
  \url{https://nlpers.blogspot.com/2015/09/how-longll-it-take-to-say-that.html}.},
which is trained on MaryTTS \cite{marytts} timing output for the most frequent German words.
In addition, we extract the position of the word in its sentence as
well as the sentence length.
For the prosodic features, we use the \textsc{snack}
library to extract power (in dB) and pitch (in semitones normalized
relative to the speaker's mean) and record the mean value of each for
each word.  The duration is extracted from the alignment.
Finally, for every word, we record whether it is followed
by a pause according to the alignment information, and if so, the
length of the pause (or zero otherwise).

\section{Experiment and Analysis} 
\label{sec:experiment}

Our experiments can be categorized into two parts: experiments
regarding nouns and experiments regarding verbs.  
We specify each experiment
using the notation introduced in Section~\ref{sec:model}.
Our detailed results for all comparisons are shown in Table~\ref{tab:results}.

We first look at nouns and at words modifying nouns.
Nouns frequently occur in subject as well as 
in object position. 
We compare subjects to accusative
objects as these -- in contrast to other objects in German -- do not
have case markers, thus making sure that the same word forms can occur in both
functions.

\textbf{subj\cmp obja}\; 
In Figure~\ref{fig:idea-a}, this would be ``cat''\texttildelow``dog''
(keep in mind that we  mitigating effects of nouns being more/less frequent in one of the positions
by fitting the data to predicted speaking durations).
We find that subjects are spoken with significantly higher pitch 
(a fifth of a semitone higher) and we also find that subjects are spoken significantly 
longer (by about 6\,\% compared to avg.\ noun durations in the corpus).
We also find that the average signal power is slightly lower for subjects. 
We do not think this (very small) effect is due to softer speech. 
Instead we speculate that the slower speech causes this.

\textbf{det\uprarrow(subj\cmp obja)}\; Differences in pronunciation
between subjects and objects could very well carry over to the determiners
attached to subjects and objects, respectively. 
In Figure~\ref{fig:idea-a}, this would be
the determiners attached to ``cat'' and ``dog''.
Again, we limit the object position to accusative.
We find a smaller effect on duration with determiners being 
spoken more quickly before subjects. We speculate that the extra time spent on 
speaking subjects is subtracted (to some extent) from preceding determiners. 
Also, there is significantly more pausing after the determiner and, presumably, 
most often before the following noun.

\textbf{attr\uprarrow(subj\cmp obja)}\; Similarly to determiners,
attributes (i.\,e.\ adjectives) 
could be pronounced differently depending on whether they
modify a subject or an object.  In Figure~\ref{fig:idea-a}, this would be
``black'' and ``quick''. Again, we find a significantly higher pitch 
(and likewise power) for attributes of subjects (but no significant 
effect on duration). Interestingly, the effect on pitch is higher for attributes 
than for the noun itself. 

Regarding verbs, we test words filling the main verb function
(\emph{s}, for \emph{sentence}\footnote{%
The attachment label of verbs in dependency grammars
directly encodes the equivalent of 
constituency in phrase structure grammars.
}), verbs that form the head of a subordinate clause
(\emph{neb}, for \emph{Nebensatz}, Figure~\ref{fig:idea-c}), verbs that form the head of a
relative clause (\emph{rel}, Figure~\ref{fig:idea-d}), and verbs attached to auxiliary verbs (Figure~\ref{fig:idea-b}).

\textbf{s\cmp neb}\; Of the non-main verb types, \emph{neb} exhibits
the least divergence from main verbs.  This fact is not surprising
since subordinate clauses are structurally the most similar to main
sentences. However, we find that all of \emph{neb}, \emph{rel}, and \emph{aux}
are spoken substantially longer than main verbs.

\textbf{s\cmp rel}\; In contrast to subordinate clauses, relative
clauses modify nouns, e.\,g.\ in Figure~\ref{fig:idea-d}, ``makes''
modifies ``cat''.  As relative clauses tend to be interjected into
sentences, they need to be distinguishable from the main sentence 
(and it may be efficient to mark this prosodically).
Verbs heading a relative clause are, on average, spoken with lower
pitch, less power, and lengthened, i.\,e.\ less pronounced.  This could be a strategy
to distinguish the additional information from the main content of the sentence.
\emph{rel} and \emph{aux} have a significant tendency
for pausing to follow.

\textbf{s\cmp aux}\; Auxiliary verbs express tense or other
grammatical aspects.  If an auxiliary verb is used, the main verb has
an infinite form and the auxiliary verb takes over the role of the
finite verb. In the annotation scheme we use, auxiliary verbs form the
head of clauses and sentences and the main verb is attached to it (see
Figure~\ref{fig:idea-b}).  Therefore, in this setting we compare
finite verbs with infinite ones that are combined with a finite
auxiliary verb. 
We speculate that pauses are less frequent after a main verb which may also 
have an influence on the duration.

\textbf{aux\uprarrow(s\cmp neb)}\; shows no significant results:
we cannot show that infinite verbs that co-occur with auxiliary verbs 
differ depending on whether they are embedded 
into a main sentence or subordinate positions. Given that subordinate clauses 
are structurally most similar to main sentences, this appears plausible.

\textbf{aux\uprarrow(s\cmp rel)}\; Verbs in conjunction with an
auxiliary verb behave differently in main clauses compared to relative
clauses. Similar to verbs without an auxiliary, the pitch is lower in
a relative clause.  The duration of verbs followed by a finite verb is
also lower in relative clauses than in main clauses.

\textbf{aux\uprarrow(s\cmp aux)}\; Verbs can form a chain in German,
such as ``gegessen haben müssen'', \emph{have to have eaten,
  \emph{lit.}\ eaten have must}.  If a verb is deeply nested like this, i.\,e.\
at least two verbs in the verb chain will still follow, it is spoken
with significantly lower pitch than a word not nested as deeply.

\section{Discussion}
\label{sec:discussion}

We have performed, to the best of our knowledge, the largest scale 
analysis of the correlation of syntax and prosody in terms of the amount of 
data (31,803 sentences in 46 hours of speech read by 300 speakers).

We have limited our correlation analysis to a few 
linguistically selected comparisons 
and often find significant differences between the contrasted conditions. 
Given the small number of comparisons we have performed, we have not 
corrected for multiple comparisons. 
Most results would stand even after (very conservative) Bonferroni correction, 
given very strong significances obtained by the large number of data points in our study.

In particular, we find strong effects of subject/object function on prosodic 
features, most prominently pitch and duration. 
We also find highly significant differences between the functions of verbs in 
different types of clauses (similarly to \cite{lelandais:hal-01329167}).
While overall the effect sizes may seem small, most are well above the 
\emph{just noticeable differences} for pitch \cite{nooteboom1997} and 
tempo \cite{quene2007jndtempo}. 
Some smaller effects might not be consciously `noticeable',
but could still subconsciously help disambiguation.

Using a corpus of read speech allowed us to infer syntactic structure 
using an automated parser trained for parsing factual texts.
The occasional errors of the parser, and hence the syntactic annotations that 
we use are unlikely to have yielded false positive results. In contrast, given 
that errors tend to add random noise, they should make our estimates more conservative.
However, parsers are known to be relatively weak in resolving attachment
ambiguity. This is why we did not investigate these in our study. 
In the future, we would like to investigate whether the prosody that relates to 
such ambiguities (e.\,g.\ for PP-attachments) naturally falls into clusters and 
whether these clusters overlap with `ground truth' for the attachment. This 
would allow us to investigate in detail the prosodic means by which speakers
communicate attachment information (and would help to improve parsers accordingly).
Spontaneous speech, with its richer interactional behaviours, might yield
different results (e.\,g.\ marking syntax might be less relevant as the expectations 
for `correct' syntax might be lower). In any case, automatic syntactic 
parsing for spontaneous spoken language is much harder to come by.

Our experiments were performed on a single language.  Using
Universal Dependencies \cite{NIVRE16.348} as a syntactic annotation
schema would enable to research similarities and differences in the
relation between prosody and syntax across languages, using the
different languages contained in the Spoken Wikipedia.


We have looked at very basic acoustic/auditory features in isolation, 
rather than looking at more higher-level prosodic features such as `prominence' 
or `phrasing' which themselves are constituted by a complex interplay of the 
basic auditory features. 
Our statistical analysis is hence conservative and points out only the most 
direct and most relevant correlations.
Feature combinations instead of simple flat representations
have lead to a break-through in parsing \cite{N18-1007}. 
Hence, we expect many more and more complex interplays in the 
syntax--prosody interface to be found in future work building on more complex 
notions of prosodic and syntactic features. 

There are many factors that influence prosody and that we have not modeled in our study, 
putting our results even further on the conservative side. One such aspect could 
be the information structure and 
modeling that would potentially yield even stronger results. 
On the other hand, we could also simply apply the same methodology as outlined 
here to investigate the influence of information structure (e.\,g.\ givenness) 
on prosodic properties. 
Using the Spoken Wikipedia for this research comes with the 
additional benefit that tools for extracting e.\,g.\ coreference are readily available
for Wikipedia-style texts.

\ \ 

\noindent
\textbf{Acknowledgements:} The authors would like to thank countless members 
of the Wikipedia community for providing the written and spoken material used 
as the basis of this study. We thank the anonymous reviewers for their helpful 
comments.

\bibliographystyle{IEEEtran}

\bibliography{synpro}

\end{document}